# Mobile big data analysis with machine learning

Jiyang Xie[1], Zeyu Song[1], Yupeng Li[2], and Zhanyu Ma[1]

[1] Pattern Recognition and Intelligent Systems Lab., Beijing University of Posts and Telecommunications, Beijing, China

[2] State Key Lab. of Networking and Switching Technology, Beijing University of Posts and Telecommunications, Beijing, China

**ABSTRACT**

This paper investigates to identify the requirement and the development of machine learning-based mobile big data analysis through discussing the insights of challenges in the mobile big data (MBD). Furthermore, it reviews the state-of-the-art applications of data analysis in the area of MBD. Firstly, we introduce the development of MBD. Secondly, the frequently adopted methods of data analysis are reviewed. Three typical applications of MBD analysis, namely wireless channel modeling, human online and offline behavior analysis, and speech recognition in the internet of vehicles, are introduced respectively. Finally, we summarize the main challenges and future development directions of mobile big data analysis.

**KEYWORDS**

Machine learning; mobile big data; wireless channel modeling; human behavior analysis; speech recognition

## 1. INTRODUCTION

With the success of wireless local access network (WLAN) technology (a.k.a. Wi-Fi) and the second/third/fourth generation (2/3/4G) mobile network, the number of mobile phones, which is 7.74 billion, 103.5 per 100 inhabitants all over the world in 2017, is rising dramatically [1]. Nowadays, mobile phone can not only send voice and text messages, but also easily and conveniently access the Internet which has been recognized as the most revolutionary development of Mobile Internet (M-Internet). Meanwhile, worldwide active mobile-broadband subscriptions in 2017 have increased to 4.22 billion, which is 9.21% higher than that in 2016 [1]. Figure 1 shows the numbers of mobile-cellular telephone and active mobile-broadband subscriptions of the world and main districts from 2010 to 2017. The numbers which are up to the bars are the mobile-cellular telephone or active mobile-broadband subscriptions (million) in the world of the year which increase each year. Under the M-Internet, various kinds of content (image, voice, video, etc.) can be sent and received everywhere and the related applications emerge to satisfy people's requirements, including working, study, daily life, entertainment, education, healthcare, etc. In China, mobile applications giants, i.e., Baidu, Alibaba and Tencent, held 78% of M-Internet online time per day in App which was about 2,412 minutes in 2017 [2]. This figure indicates that M-Internet has entered a rapidly growth stage.

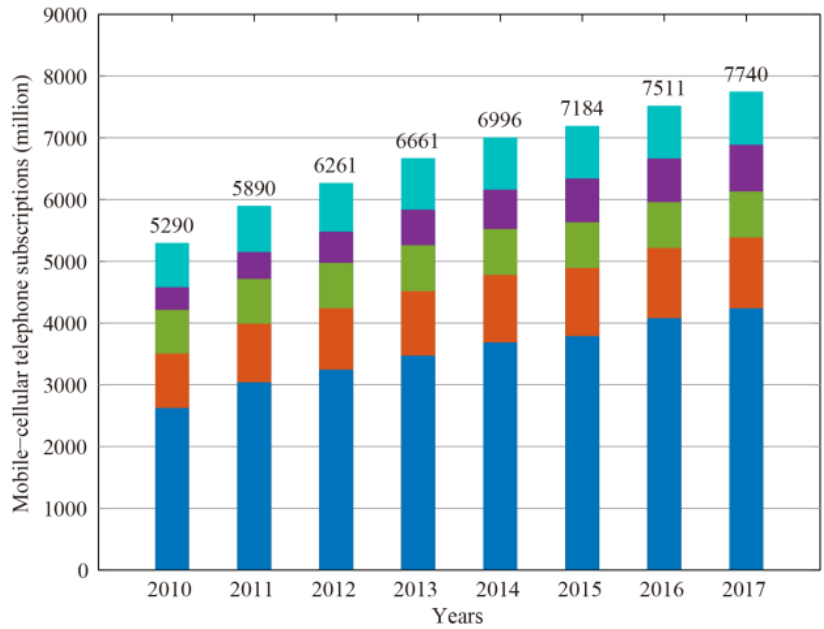

(a)

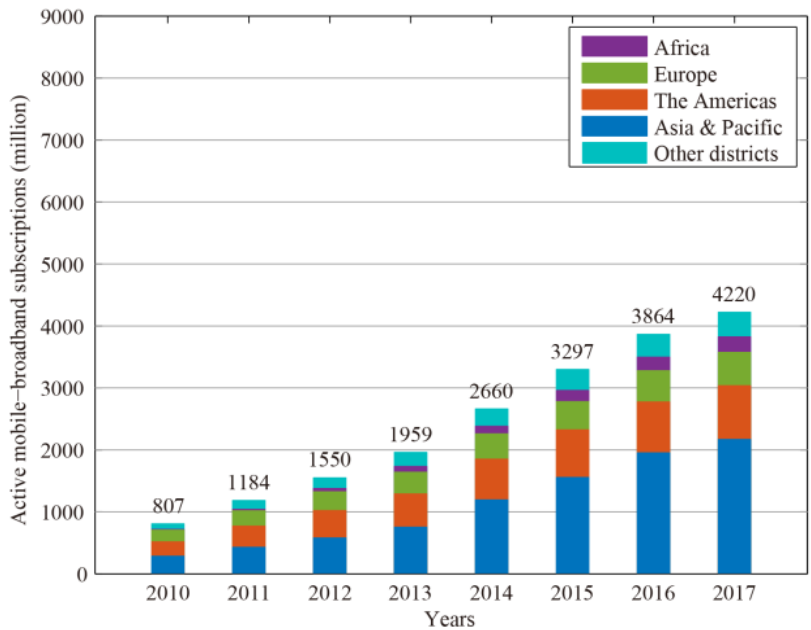

(b)

Fig.1 Mobile-cellular telephone subscriptions (million) in (a) and active mobile-broadband subscriptions (million) in (b) of the world and main districts [1].

Nowadays, more than one billion smart phones are in use that are producing vast amount of data. This is bringing profound impact on society and social interaction as well as increasing tremendous opportunities for business. Meanwhile, with the rapid development of the Internet-of-Things (IoT), much more data is automatically generated by millions of machine nodes with growing mobility, e.g., sensors carried by moving objects or vehicles. The volume, velocity, and variety of this data are increasing extremely fast, and soon it will become the new norm for enterprise analytics. Therefore, MBD has been already in our lives and being enriched rapidly. The trend for explosively increased data volume with the increasing bandwidth and data rate in the M-Internet has followed the same exponential increase as Moore's Law for semiconductors [3]. The prediction [2] about the global data volume will grow up to 47 zettabytes (1 zettabyte = $1 \times 10^{21}$ bytes) by 2020 and 163 zettabytes by 2025. For M-Internet, 3.7 exabytes (1 exabyte = $1 \times 10^{18}$ bytes) data have been generated per month from the mobile data traffic in 2015 [4], 7.2 exabytes in 2016 [5], 24 exabytes by 2019 on forecasting [6], and 49 exabytes by 2021 on forecasting [5]. According to the statistical and prediction results, a concept called Mobile big data (MBD) has appeared.

The MBD is a concept of a massive amount of mobile data which are generated from a large amount of mobile devices and cannot be processed using a single machine [7], [8]. MBD is playing and will play a more important role than ever before by the popularization of mobile devices including smartphones and IoT gadgets especially in the era of 4G and the forthcoming 5G [4], [9].

With the rapid development of information technology, various kinds of data from different industry are showing explosive growth trends [10]. Big data has broad application prospects in many fields, and has become important national strategic resources [11]. In the era of big data, many data analysis systems are facing big challenges as the volume of data increases. Therefore, analysis for MBD is currently a highly focused topic. The importance of MBD analysis is determined by its role in building complex mobile systems which support many intelligently interactive services, for example, healthcare, smart building, and online games [4]. MDB analysis can be defined as mining terabyte-level or petabyte-level data collected from mobile users and wireless devices at the network-level or the app-level to discover unknown, latent and meaningful patterns and knowledge with large scale machine learning methods [12].

Present requirements of MBD are based on software-defined in order to be more scalable and flexible. M-Internet environment in the future will be even more complex and interconnected [13]. For this purpose, data centers of MBD need to collect user statistics information of millions of users and obtain meaningful results by proper MBD analysis methods. For the decreasing price of data storage and widely accessible high performance computers, an expansion of machine learning has come into not only theoretical researches, but also various application areas of big data. Even though, there is a long way to go for the machine learning-based MBD analysis.

Machine-learning technology powers many aspects of modern society: from web searches [14], [15] to content filtering on social networks [16] to recommendations on e-commerce websites [17], [18]. Furthermore, it is also frequently appearing in consumer products such as cameras and smartphones. Machine-learning systems are used to identify objects in images, transcribe speech into text, match news items, posts or products with users' interests, and select relevant results of search. In recent years, big data machine learning has become a hot spot [19]. Some conventional machine learning methods based on Bayesian framework [20], [21], [22], [23] and distributed optimization [24], [25], [26], [27] can be applied into the aforementioned applications and have obtained good performances in small data sets. On this foundation, researchers have always been trying to fill their machine learning model with more and more data [28]. Furthermore, the data we got is not only big but also has features such as multi-source, dynamic, sparse value etc., these features make it harder to analyze MBD with conventional machine learning methods. Therefore, the aforementioned applications implemented with conventional machine learning methods have fallen in a bottleneck period for low accuracy and generalization. Recently, a class of novel techniques, called deep learning, is applied in order to make the effort to solve the problems, and has obtained good performances [29]. Machine learning, especially deep learning, has been an essential technique in order to use big data effectively.

In contrast to most conventional learning methods, which are considered using shallow-structured learning architectures, deep learning refers to machine learning techniques that use supervised and/or unsupervised strategies to automatically learn hierarchical representations in deep architectures for regression and classification [30], [31]. Deep learning algorithms are one promising avenue of research into the automated extraction of complex data representations (features) at high levels of abstraction using a huge amount of unsupervised data to automatically extract complex representation [32], [33]. Such algorithms develop a layered, hierarchical architecture of learning and representing data, where higher-level (more abstract) features are defined in terms of lower-level (less abstract) features. Empirical studies have demonstrated that data representations obtained from stacking up nonlinear feature extractors (as in deep learning) often yield better machine learning results, e.g. improved classification modeling [34], better quality of generated samples by generative probabilistic models [35], and the invariant property of data representations [36]. Deep learning solutions have yielded outstanding results in different machine learning applications. A more detailed overview of deep learning is presented in Section 3.1.

Artificial intelligence (AI), which has the general goal of emulating the human brain's ability to observe, analyze, learn, and make decisions, especially for extremely complex problems, motivates the development of deep learning algorithms. Work pertaining to these complex challenges has been a key motivation behind deep learning algorithms which strive to emulate the hierarchical learning approach of the human brain. Models based on shallow learning architectures such as decision trees, support vector machines, and case-based reasoning may fall short when attempting to extract useful information from complex structures and relationships in the input corpus. In contrast, inspired by biological observations on human brain mechanisms for processing of natural signals [37], [38], [39], deep learning architectures have the capability to generalize in non-local and global ways, generating learning patterns and relationships beyond immediate neighbors in the data, and have attracted much attention from the academic community in recent years due to its state-of-the-art performance in many research domains such as not only the aforementioned natural language processing (NLP), but also speech recognition [40], [41], collaborative filtering [42], and computer vision [43], [44].

Deep learning has also been successfully applied in industry products that take advantage of the large volume of mobile data. Companies like Google, Apple, and Facebook, which collect and analyze massive amounts of mobile data on a daily basis, have been aggressively pushing forward deep learning related projects. For example, Apple's Siri, the virtual personal assistant in iPhones, offers a wide variety of services including weather reports, sport news, answers to user's questions, and reminders etc. by utilizing deep learning and more and more data collected by Apple services [45]. Google applies deep learning algorithms to massive chunks of messy data obtained from the Internet for Google's translator.

MBD contains a large variety of information of offline data and online real-time data stream generated from smart mobile terminals, sensors and services, and hastens various applications based on the advancement of data analysis technologies, such as, collaborative filtering-based recommendation [46], [47], user social behavior characteristics analysis [48], [49], [50], [51], vehicle communications in the Internet of Vehicles (IoV) [52], online smart healthcare [53], and city residents' activity analysis [7]. Although the machine learning-based methods are widely applied in the MBD fields, and obtain good performances in real data test, the present methods still need to be further developed. Therefore, five main challenges facing MBD analysis regarding the machine learning-based methods include large-scale and high-speed M-Internet, overfitting and underfitting problems, generalization problem, cross-modal learning and extended channel dimensions and should be considered.

This paper investigates to identify the requirement and the development of machine learning-based mobile big data analysis through discussing the insights of challenges in the MBD and reviewing state-of-the-art applications of data analysis in the area of MBD. The remainder of the paper is organized as follows: Section 2 introduces the development of data collection and properties of MBD. The frequently adopted methods of data analysis and typical applications are reviewed in Section 3. Section 4 summarizes the future challenges of MBD analysis, and provides suggestions.

## 2. DEVELOPMENT AND COLLECTION OF THE MOBILE BIG DATA
### 2.1. Data collection

Data collection is the foundation of a data processing and analysis system. Data are collected from mobile smart terminals and Internet services, or called mobile Internet devices (MIDs) generally, which are multimedia-capable mobile devices providing wireless Internet access and contain smartphones, wearable computers, laptop computers, wireless sensors, etc. [54]. Mobile big data should be collected with fast and accurate manner in a non-disturbing way [8].

MBD can be divided into two hierarchical data form: transmission and application data, from bottom to top. The transmission data focus on solving channel modeling [55], [56] and user access problems corresponding to the physical transmission system of M-Internet. On this foundation, application data focus on the applications based on the MBD including social networks analysis [57], [58], [59], user behavior analysis [48], [50], [60], speech analysis and decision in IoV [61], [62], [63], [64], [65], [66], smart grid [67], networked healthcare [53], [68], [69], Finance services [46], [70], etc.

Due to the heterogeneity of the M-Internet and the variety of the access devices, the collected data are unstructured and usually in various categories and formats, which make data pre-processing become an essential part of a data processing and analysis system in order to ensure the input data complete and reliable [71]. Pre-processing can be usually divided into three steps which are data cleaning, generation of implicit ratings and data integration [46].

**1) Data cleaning**

Due to possible equipment failures, transmission errors or human factor, raw data are "dirty data" which cannot be directly used, generally [46]. Therefore, data cleaning methods including outlier detection and denoising are applied in the data pre-processing to obtain the data meet required quality. Manual removal of error data is difficult and impossible to accomplish in MBD due to the massive volume. Common data cleaning methods can alleviate the dirty data problem to some extent by training support vector regression (SVR) classifiers [72], multiple linear regression models [73], autoencoder [74], Bayesian methods [75], [76], clustering models, distance-based models, density-based models, probabilistic models or information-theoretic models [77].

**2) Generation of implicit ratings**

Generation of implicit ratings is mainly applied in recommend systems. The volume of rating data can be increased greatly by analysing specific user behaviors to solve data sparsity problem with machine learning algorithms, for example, neural networks and decision trees [46].

**3) Data integration**

Data integration is a step to integrate data from different resources with different formats and categories, and to handle missing data fields [8].

Figure 2 represents the procedures of data collection and pre-processing.

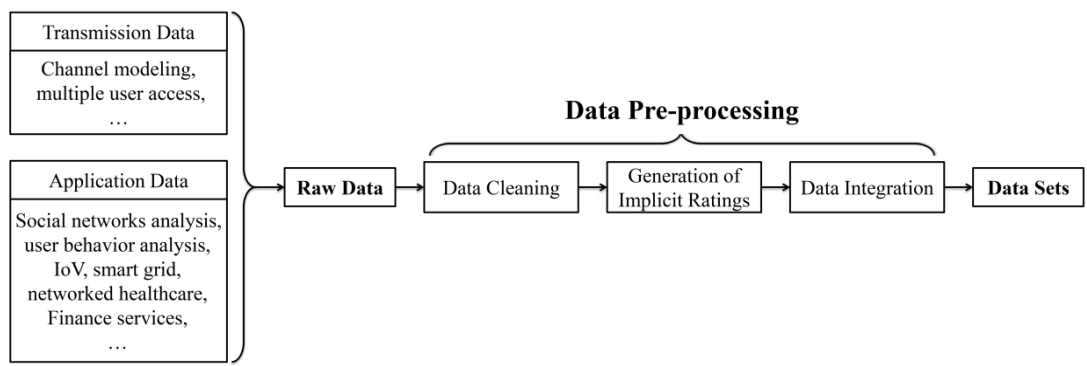

Fig. 2 The procedures of data collection and pre-processing.

**2.2. Properties of mobile big data**

The MBD brings many new challenges to conventional data analysis because of its high dimensionality, heterogeneity, and other complex features from the applications, such as planning, operations and maintenance, optimization, and marketing [57]. This section discusses the five Vs (short for volume, velocity, variety, value and veracity) features [78] deriving from big data towards the MBD. The five Vs features have been enhanced in M-Internet causing it enable users to access Internet anytime and anywhere [79].

1) **Volume**: large number of MIDs, Exabyte-level data and high-dimensional data space

Volume is the most obvious feature of MBD. In the forthcoming 5G network and the era of MBD, conventional store and analysis methods are incapable of processing the 1000x or more wireless traffic volume [8], [80]. It is of great urgency to improve present MBD analysis methods and propose new ones. The methods should be simple and cost-effective to be implemented for MBD processing and analysis. Moreover, they should also be effective enough without requiring large amount of data for model training. And finally, they are precise to be applied in various fields, [79].

2) **Velocity**: real-time data streams and efficiency requirement

Velocity addresses the speed at which data can be received as well as analyzed [81]. The data is now continuously streaming into the servers in real time, and makes the original batch process break down [82]. Due to the high generating rate of MBD, velocity is the efficiency requirement of MBD analysis since real-time data processing and analysis are extremely important in order to maximize the value of MBD streams [8].

3) **Variety**: heterogeneity and non-structured of mobile multimedia contents

Due to the heterogeneity of MBD which means that data traffic comes from spatially distributed data resources (i.e. MID), the variety of MBD arises and makes the MBD more complex [4]. Meanwhile, the non-structured of MBD also causes the variety. The MBD can be divided into structured data, semi-structured data and unstructured data. Unstructured data are often collected in new applications and have random data fields and contents [8], therefore, they are difficult to analyze before data cleaning and integration.

4) **Value**: mining hidden knowledge and patterns from low density value data

Value, or called low density value of MBD, is caused by a large amount of useless or repeated information in the MBD. Therefore, we need to mine the big value by MBD analyzing which is hidden knowledge and patterns extraction. The purified data can provide a wide range of information to conduct more effectively analysis about user demand characteristics, user behavior and user habits [83], and to achieve better management, more accurate prediction and decision-making [84].

**5) Veracity**: consistency, trustworthiness and security of MBD

The veracity of MBD includes two parts: data consistency and trustworthiness [78]. It can also be summarized as data quality. MBD quality is not guaranteed due to the noise of transmission channel, the equipment malfunctioning, the uncalibrated sensors of MIDs or the human factor (for instance, malicious invasion) resulting in low-quality data points [4]. MBD veracity ensures that the data used are trusted, authentic and protected from unauthorized access and modification where the data must be collected from trusted sources relying entirely on the security infrastructure deployed and available from the MBD infrastructure [78].

## 3. APPLICATIONS OF MACHINE LEARNING METHODS IN THE MOBILE BIG DATA ANALYSIS

### 3.1. Development of data analysis methods

In this section, we present some recent achievements in data analysis from four different perspectives.

### 3.1.1. Divide-and-conquer strategy and sampling of big data

Divide and conquer strategy is a computing paradigm dealing with big data problems. After the development of distributed and parallel computing, the divide and conquer strategy is particularly important.

Generally speaking, the importance of different samples in learning data is also different. Some redundant and noisy data can not only cause a large amount of storage cost, but also reduce the efficiency of the learning algorithm and affect the learning accuracy. Therefore, it is more preferable to select representative samples to form a subset of the original sample space according to a certain performance standard, such as maintaining the distribution of samples, topological structure and keeping classification accuracy. Then learning method will be constructed on this subset to complete the learning task. In this way, we can maintain or even improve the performance of big data analyzing algorithm with minimum computing and stock resources. Under the background of big data, the demand for sample selection is more urgent. But most of the sample selection method is only suitable for smaller data sets, such as the traditional Condensed Nearest Neighbor (CNN) [85], the Reduced Nearest Neighbor (RNN) [86] and the Edited Nearest Neighbor (ENN) [87], the core concept of these methods is to find the minimum consistent subset. To find the minimum consistent subset, we need to test every sample and the result is very sensitive to the initialization of the subset and samples setting order. Li et al. [88] proposed a method to select the classification and edge boundary samples based on local geometry and probability distribution. They keep the space information of the original data, but need to calculate k-means for each sample. Angiulli et al. [89], [90] proposed a fast nearest neighbor algorithm (FCNN) based on CNN, which tends to choose the classification boundary samples.

Jordan [91] proposed statistical inference method for big data. When dealing with statistical inference with divide and conquer algorithm, we need to get confidence intervals from huge data sets. The Bootstrap theory aims to obtain the fluctuation of the evaluation value by re-sampling the data and then obtain the confidence interval, but it is not feasible for big data. The incomplete sampling of data can lead to erroneous range fluctuations. It must be corrected in order to provide statistical inference calibration. An algorithm named Bag of Little Bootstraps was proposed, which can not only avoid this problem, but also has many advantages on computation. Another problem discussed in [91] is massive matrix calculation. The divide and conquer strategy is heuristic, which has a good effect in practical application. However, new theoretical problems rise when trying to describe the statistical properties of partition algorithm. To this end, the support concentration theorem based on the theory of random matrices has been proposed.

As a conclusion, data partition and parallel processing strategy is the basic strategy to deal with big data. But the current partition and parallel processing strategy uses little data distribution knowledge, which has influence the load balancing and the calculation efficiency of big data processing. Hence, there exists an urgent requirement to solve the problem about how to learn the distribution of big data for the optimization of load balancing.

### 3.1.2. Feature selection of big data

In the field of data mining, document classification and indexing in multimedia data, the dataset is always large, which contains a large number of records and features. This leads to the low efficiency of algorithm. By feature selection, we can eliminate the irrelevant features and increase the effectiveness of task analysis. Thus we can improve the accuracy of the model and reduce the running time.

A huge challenge of big data processing is how to deal with the high dimensional and sparse data. Under big data environment, network traffic, communication records, large-scale social network have a large number of high-dimensional data, the tensor (such as a multidimensional array) representation provides a natural representation of data. Tensor decomposition, in this condition, becomes an important tool for summary and analysis. Kolda [92] proposed an efficient use of the memory of the Tucker decomposition method (Memory - Efficient Tucker Decomposition, MET) to solve the time and space usage problem which can't be solved by traditional tensor decomposition algorithm. MET adaptively selects execution strategy based on available memory in the process of decomposition. The algorithm maximizes the speed of computation in the premise of using the available memory. MET avoid dealing with the large number of sporadic intermediate results proceeded during the calculation process. The adaptive selections of operation sequence, not only eliminate the intermediate overflow problem, but also save memory without reducing the precision. On the other hand, Wahba [93] proposed two approaches to the statistical / machine learning model which involve discrete, noisy and incomplete data. These two methods are Regularized Kernel, Estimation (RKE) and Robust Manifold robust manifold Unfolding (RMU). These methods use the dissimilarity between training information to get a non-negative low rank definite matrix. The object will be embedded into a low dimensional Euclidean space, which coordinate can be used as features of various learning modes. Similarly, most online learning research needs to access all features of training instances. Such classic scenario is not always suitable for practical applications when facing high-dimensional data instances or expensive feature sets. In order to break through this limit, Hoi et al. [94] propose an efficient algorithm to predict online features solving problem using a small and fixed number of active features based on their study of sparse regularization and truncation technique. They also evaluate the proposed algorithm in some public data sets in line feature selection and performance experience.

The traditional self-organizing map (SOM) can be used for feature extraction. But when the data set is large, SOM has the shortcoming of slow speed. Sagheer [95] proposed a fast self-organizing map (FSOM) to solve this problem. The main idea of this method is: a main area the information data is mainly distributed in the feature space, if we can find these areas and directly extract information in these areas, rather than the information extraction in the whole data space, it can greatly reduce the time.

Anaraki [96] proposed a threshold method of fuzzy rough set feature selection based on fuzzy lower approximation. This method adds a threshold to limit the QuickReduct feature selection. The results of the experiment prove that this method can improve the accuracy of feature extraction and reduce running time.

Gheyas et al. [97] proposed a hybrid algorithm of simulated annealing and genetic algorithm (SAGA), combining the advantages of simulated annealing algorithm, genetic algorithm, greedy algorithm and neural network algorithm, to solve the NP-hard problem of selecting optimal feature subset. The experiment shows that this algorithm can find better optimal feature subset, and can also reduce the time complexity. Gheyas pointed out at the end of the paper that there is seldom a single algorithm which can solve all the problems, the combination of algorithms can effectively raise the overall affect.

To sum up, because of the complexity, high dimensionality and uncertain characteristics of big data, it is an urgent problem to solve how to reduce the difficulty of big data processing by using dimension reduction and feature selection technology.

### 3.1.3. Big data classification

Supervised learning (classification) faces a new challenge of how to deal with big data. Currently, classification problems involving large scale data are ubiquitous, but the traditional classification algorithms do not fit big data processing properly.

### 3.1.3.1. SVM classification

Traditional statistical machine learning method has two main problems when facing big data. 1) Traditional statistical machine learning methods are always involving intensive computing which makes it hard to apply on big data sets. 2) The prediction of model that fits the robust and non-parameter confidence interval is unknown. Lau et al. [98] proposed an online SVM learning algorithm to deal with the classification problem for sequentially provided input data. The classification algorithm is faster, with less number of support vectors, and has better generalization ability. Laskov et al. [99] proposed a rapid, stable and robust numerical incremental support vector machine learning method.

In addition, Huang et al. [100] presents a large margin classifier M4. Unlike other large margin classifiers which locally or globally constructed separation hyperplane, this model can learn both local and global decision boundary. SVM and minimax probability machine (MPM) has a close connection with the model. The model has important theoretical significance and furthermore, the optimization problem of M4 can be solved in polynomial time.

### 3.1.3.2. Decision tree classification

Traditional decision tree, as a classic classification learning algorithm, has a large memory requirement problem when processing big data. Franco-Arcega et al. [101] put forward a method of constructing decision tree from big data, which overcomes some limitations of current algorithm. Furthermore, it can use all training set data without saving them in memory. Experimental results showed that the method is faster than the current decision tree algorithm on large scale problems. Yang et al. [102] proposed a fast incremental optimization decision tree algorithm for large data processing with noise. Compared with the traditional decision tree data mining algorithm, the main advantage of this algorithm is real-time mining ability, which is quite suitable when the mobile data stream is unlimited. Moreover, it can store the complete data for training decision model. The advantage of this model is that it can prevent the explosive growth of the decision tree size and the decrease of prediction accuracy when the data packet contains noise. The model can generate compact decision tree and predict accuracy even with highly noisy data. Ben-Haim et al. [103] proposed an algorithm of building parallel decision tree classifier. The algorithm runs in distributed environment and is suitable for large data sets and streaming data. Compared with serial decision tree, the algorithm can improve efficiency under the premise of accuracy error approximation.

### 3.1.3.3. Neural network and extreme learning machine

Traditional feed-forward neural networks usually use gradient descent algorithm to tune the weight parameters. Generally speaking, slow learning speed and poor generalization performance are the bottlenecks that restrict the application of feed-forward neural network. Huang et al. [104] discarded the iterative adjustment strategy of the gradient descent algorithm and proposed extreme learning machine (ELM). This method randomly assigns the input weights and the deviations of the single hidden layer neural network. It can analyze the output weights of the network by one step calculation. Compared to the traditional feedforward neural network training algorithm, the network weights can be determined by multiple iterations, and the training speed of ELM is significantly improved.

However, due to the limitation of computing resource and computational complexity, it is a difficult problem to train a single ELM on big data. There are usually two ways to solve this problem: 1) training ELM [105] based with divide and conquer strategy; 2) introducing parallel mechanism [106] to train a single ELM. It is shown in references [107], [108] that a single ELM has strong function approximation ability. Whether it is possible to extend this approximation capability to ELM based on divide and conquer strategy is a key index to evaluate the possibility that ELM can be applied to big data. Some of the related studies also include effective learning to solve such problem [109].

In summary, the traditional classification method of machine learning is difficult to apply to the analysis of big data directly. The study of parallel or improved strategies of different classification algorithms has become the new direction.

### 3.1.4. Big data deep learning

With the unprecedentedly large and rapidly growing volumes of data, it is hard for us to get hidden information from big data with ordinary machine learning methods. The shallow-structured learning architectures of most conventional learning methods are not fit for the complex structures and relationships in these input data. Big data deep learning algorithm, with its deep architectures, is able to generalize in non-local and global ways, learning complex patterns and relationships beyond big data [37], [110]. It has had state-of-art performances in many research domains and also been applied in industry products to deal with large amount of digital data. In this section, we will introduce some deep learning methods in big data analytics.

Big data deep learning has some problems: 1) The deep nets have many hidden layers that make it difficult to learn when given a data vector, 2) the need of all the parameters to be learned together makes the learning time increasing sharply as the number of parameters arises, 3) the approximations at the deepest hidden layer may be poor. Hinton et al. [32] proposed a deep architecture: deep belief network (DBN) which can learn from both labeled and unlabeled data by using unsupervised pre-training method to learn data distributions without label and a supervised fine-tune method to construct the models, and solved part of the aforementioned problems. Meanwhile, subsequent researches, for example, [111], improved the DBN trying to solve the problems.

Convolutional neural network (CNN) [112] is another popular deep learning network structure for big data analyzing. A CNN has three common features including local receptive fields, shared weights and spatial or temporal sub-sampling, and two types of layers called convolutional and subsampling layers [113], [114]. CNN is mainly applied in computer vision field for big data, for example, image classification [115], [116] and image segmentation [117].

Document (or textual) representation, or called natural language processing (NLP), is a key aspect in information retrieval for many domains. The goal of document representation is to create a representation that condenses specific and unique aspects of the document, e.g. document topic. In practice, it is often observed that the occurrences of words are highly correlated. Using deep learning techniques to extract meaningful data representations makes it possible to obtain semantic features from such high-dimensional textual data, which in turn also leads to the reduction of the dimensions of the document data representations. Hinton et al. [118] proposed a deep learning generative model to learn the binary codes for documents which can be used for information retrieval and easy to store up. Socher et al. [119] proposed a recursive neural network on parsing natural scenes and natural language processing, and achieved state-of-art results on segmentation and annotation of complex scenes. Kumar et al. [120] proposed that recurrent neural networks can construct a search space from large amount of big data.

The works conducted in, for example, [40], [41], [43], [121], [122], [123], [124], proposed effective and scalable parallel algorithms for training deep models with the unprecedented growth of commercial and academic data sets in recent years. Researchers focus on large scale deep learning that can be implemented in parallel including improved optimizers [122] and new structures [112], [124], [125], [126].

In conclusion, big data deep learning methods are the key methods of data mining. They use complex structure to learn patterns from big data sets and various volumes of non-traditional data. The development of data storage and computing technology promotes the development of deep learning methods and make it easier to use in practical situations.

### 3.2. Wireless channel modeling

It is known that wireless communication conveys the information by electromagnetic waves between the transmitter antenna and receiver antenna, which is recognized as a wireless channel. In the last decades, the channel dimensions are continuously expanded from time domain to space, time and frequency three domains, which mean that the channel properties are deeply mined. Another progress is that the channel characteristics are precisely described by different methodologies, i.e., channel modeling [127].

Liang et al. [128] used machine learning to predict channel state information in order to decease the pilot overhead. Especially for 5G, the wireless big data appears and its related techniques are applied to conventional communication research in order to realize the vision in 5G. However, the wireless channel is physical electromagnetic waves in essence and currently 5G channel model researches still follow the conventional ways. Considering that the channel measurement data of 5G already appears in big volume because of the increased antenna number, huge bandwidth and versatile application scenarios, this part will introduce the interdisciplinary research of big data and wireless channel, i.e., a cluster-nuclei based channel model [129]. The novel model takes advantages of both the stochastic model and deterministic model. The channel data is collected and the channel parameters are estimated. Based on these, the multipath components (MPCs) are clustered as the conventional stochastically channel model. In parallel, the scenario is recognized by computer and the environment is reconstructed by machine learning method. Then the cluster-nuclei are found by matching the clusters with the real propagation objects, which are the key elements to bridge the deterministic environment and the stochastic clusters. Thereby, a "wave, cluster-nuclei, channel" three-layer structure is formed naturally. Finally, the channel impulse response (CIR) predication in various scenarios and configurations can be realized by cluster-nuclei based channel model with the rules explored from the channel database of various scenarios, frequencies and antenna configurations. Thus this model is promising to support the 5G and beyond research for its low complexity of limited cluster-nuclei as well as physical mapping between the clusters and the objects.

### 3.2.1. A cluster-nuclei based model

With the three-layer structure of "char, stroke, sub-stroke", the computer can learn automatically like human being by Bayesian learning program [130]. All of above tell us the truth that the basic and simple generating elements and structures are hidden in the complicated, versatile wireless propagation phenomenon. Considering that the radio waves encounter the main scatters between the transmitter and the receiver, the received signal is their synthetic effects of reflection, scattering and diffraction, which will appear as MPC clusters. So there must be the relation between the MPC clusters defined in stochastic model and the scatters in the deterministic environment.

In this structure, a cluster-nuclei is defined as one of clusters which is aggregated by a large number of waves (MPC). There are three important features for cluster-nuclei. 1) It has a certain shape. 2) It has the mapping relation between scatters in the real propagation environment and clusters. 3) It dominates the channel impulse response generation in various scenarios and configurations. With the introduction of cluster-nuclei, the three-layer structure is formed, that is, "the wave, cluster nuclei and channel". The reason to introduce cluster-nuclei is to decrease the complexity such that we can model the channel from numerous MPCs directly. Another reason is that cluster-nuclei is mapped with the environment objects, rather than the conventionally clusters only stochastically obtained from delay, angles and power.

With the proposed three-layer structure, the mapping rules between scatters and cluster nuclei can be explored and the produce method of channel impulse response can be learned by using data mining techniques and machine learning algorithms.

The MPCs are clustered with the Gaussian mixture model (GMM) [131]. Using sufficient statistic characteristics of channel multipaths, the GMM can get clusters corresponding to the multipath propagation characteristics. Figure 3 illustrates the simulation result of GMM clustering algorithm.

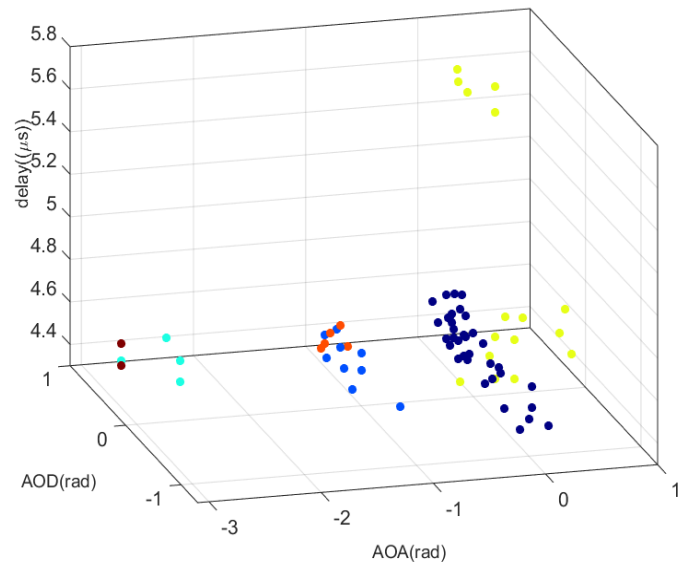

Fig. 3 Clustering results of GMM [131].

At the meantime, simultaneous localization and mapping (SLAM) algorithm is used to identify the texture from the measurement scenario picture in order to reconstruct 3 dimensional (3D) propagation environment and find the main deterministic objects [132], [133]. Figure 4 illustrates the reconstruction result with SLAM algorithm.

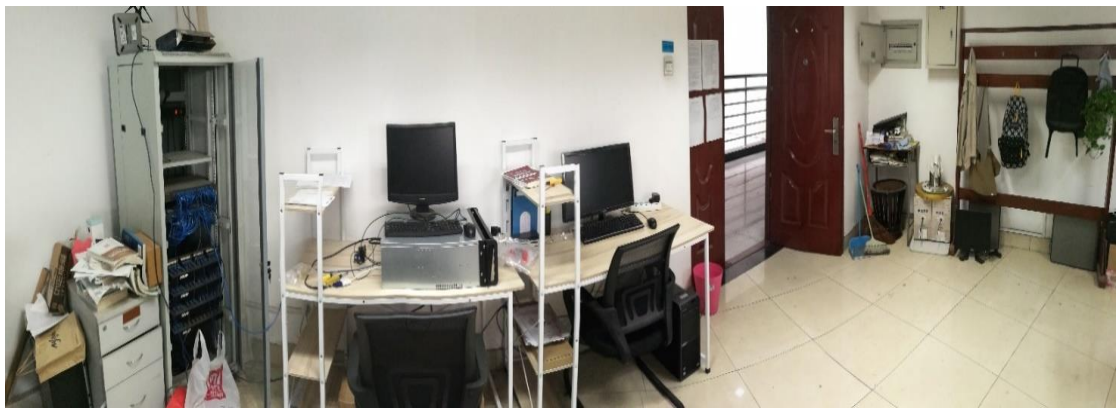

(a)

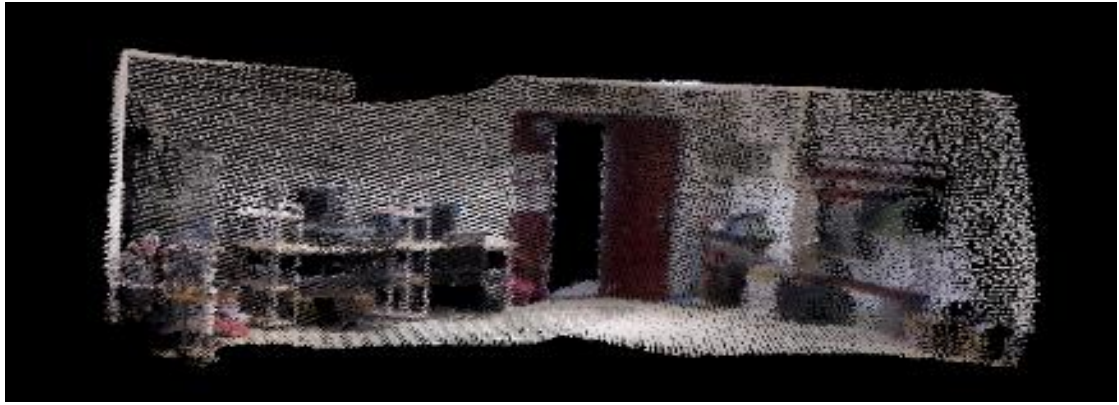

(b)

Fig. 4 Recognition of Multi-objects with SLAM algorithm: (a) real indoor scene; (b) reconstruction result with SLAM algorithm.

The key step is to search the mapping and matching relation between clusters and scatters based on the cluster characteristics and objects properties.

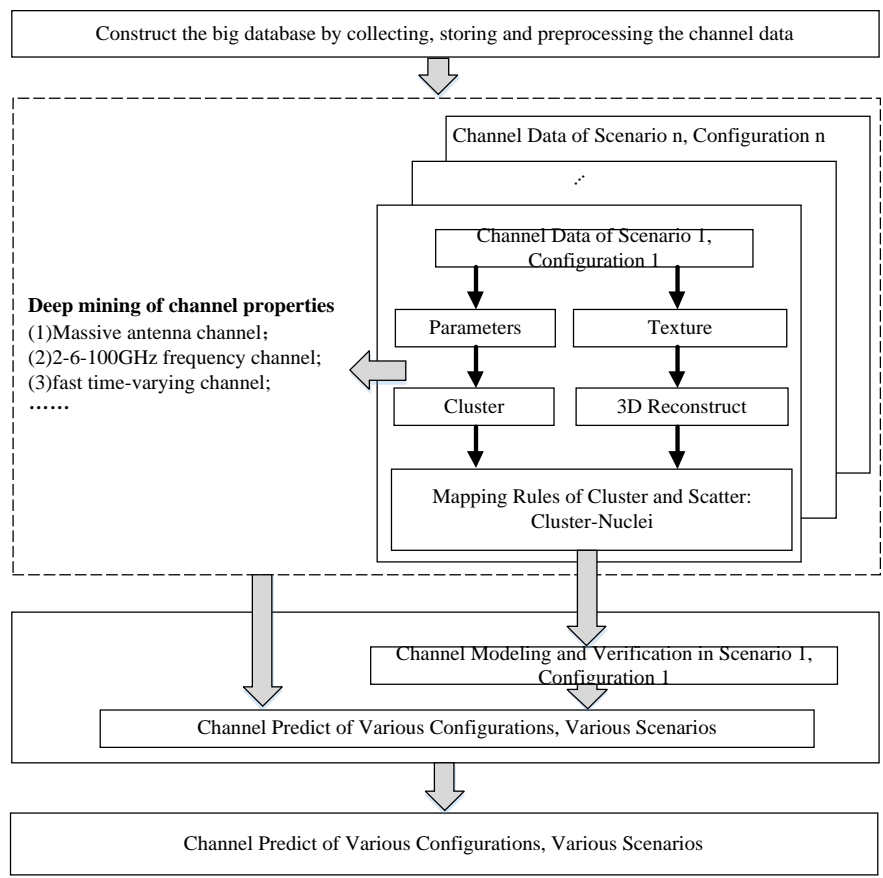

Fig. 5 The principle of the cluster-nuclei based channel modeling [129].

Then the mechanism and methods to form the cluster-nuclei is clear. With the limited number of cluster-nuclei, the channel impulse response can be produced by machine learning, i.e., decision tree [134], neural network [135]. Based on the database from various scenarios, frequency and antenna configurations, channel changing rules can be explored and found, and then input to the cluster-nuclei based modeling. Finally, the channel impulse response predication in various scenarios and configuration can be realized.

Stochastic channel models like geometry-based stochastic modeling (GBSM) lack physical meaning and deterministic channel models like ray tracing and the map-based modeling method are highly complex and rely on the precision of the geographic information. In contrast, the proposed model takes advantages of both the stochastic and deterministic models, i.e., low complexity with the limited number of cluster-nuclei while cluster-nuclei has the physical mapping to real propagation objects. In addition, channel properties are explored from database, including the variation with the antenna number, frequency, mobility and etc. In order to support massive multiple-input multiple-output (MIMO), high frequency band and high mobility required by 5G and beyond. Such properties will be input to the channel realization to achieve the channel impulse response predication in versatile environments.

**3.3. Analyses of human online and offline behavior based on mobile big data**
The advances of wireless technologies and ever-increasing mobile applications bring about explosion of mobile traffic data. It is an excellent source of knowledge to obtain the movement regularity of individuals and acquire the mobility dynamics of populations of millions [136]. Previous researches have described how individuals visit geographical locations and employed mobile data to infer human mobility patterns. Representative works like [137], [138] explore the mobility of users in terms of the number of cells they visited, which turn out to be a heavy tail distribution. Authors in [137], [139], [140] also reveal that a few important locations are frequently visited by users. Especially, these preferred locations are usually related to home and work places. Moreover, through defining a measure of entropy, Song et al. [141] believes that 93% of individual movements are potentially predictable. Thus, various models have been applied to describe the human offline mobility behavior [142]. In general, passively collecting human mobile traffic data, while he/she is accessing the mobile Internet, has lots of advantages: high cost efficiency, low energy consumption, covering a wide range and a large number of people, and with fine time granularity, which give us the opportunity to study human mobility at a scale that other data source very hard to reach [143]. The novel mobility models obtained from mobile data are expected to affect a number of fields, including urban planning, road traffic engineering, human sociology, epidemiology of infectious diseases, or telecommunication networking [136].

Online browsing behavior is another important feature on user behavior of network resource consumption. A variety of applications are now available on smart devices, covering all aspects of our daily life and providing convenience. For example, we can order taxies, shop, and book hotels using mobile phones. Yang et al. [49] provide a comprehensive understanding of user behavior in mobile Internet. It has been found that many factors, such as data usage and mobility pattern, may impact people's online behavior on mobile devices. It is discovered that the more the number of distinct cells a user visit, the more diverse applications user has visited. Zheng et al. [144] analyze the longitudinal impact of proximity density, personality, and location on smartphone traffic consumption. In particular, location has been proven to have strong influences on what kinds of apps users choose to use [140], [144]. The aforementioned observations point out that there is a close relationship between online browsing behavior and offline mobility behavior.

Figure 6(a) is an example of how browsed applications and current location related with each other from the view of temporal and spatial regularity. It has been found that, the mobility behaviors have strong influences on online browsing behavior [140], [144], [145]. Similar trends can also be observed for crowds at crowd gathering places, as is shown in Figure 6(b), i.e., certain apps are favored at places that group people together and provide some specific functions. The authors in [50] tried to measure the relationship between human mobility and app usage behavior. In particular, the authors proposed a rating framework which can forecast the online app usage behavior for individuals and crowds. Building the bridge between human mobility and mobile Internet can tell us what people need in daily activities. Content providers can leverage this knowledge to appropriately recommend content for mobile users. At the same time, Internet service providers (ISPs) can use this knowledge to optimize networks for better end-user experiences.

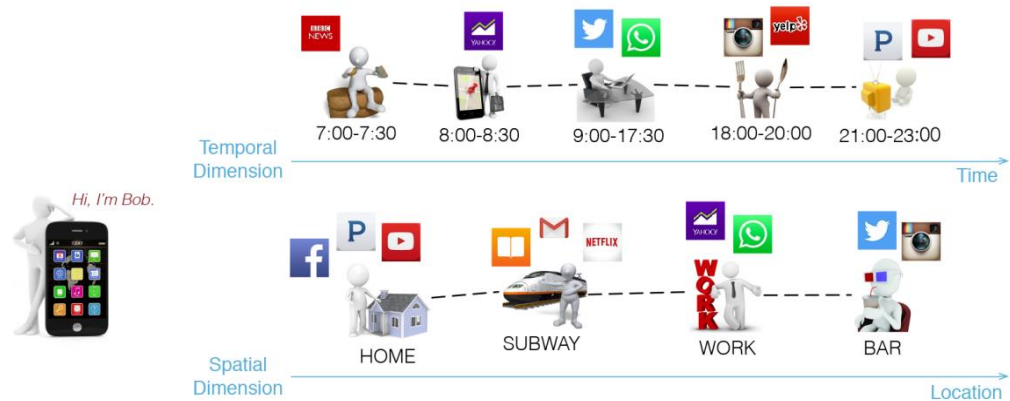

(a) App usage behavior of Bob in temporal and spatial dimension.

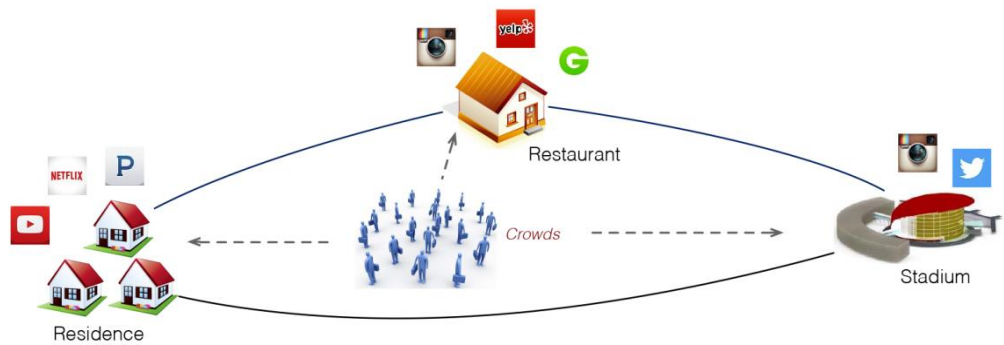

(b) App usage behavior of crowds at crowd gathering place.

Fig. 6 App usage behavior in daily life: (a) the app usage behavior of an individual; (b) app usage behavior of crowds at crowd gathering places [50].

In order to make full use of users' online and offline information, some researchers begin to quantize the interplay between online social network and offline social network and investigate network dynamics from the view of mobile traffic data [146], [147], [148], [149]. Specifically, the online and offline social networks are respectively constructed based on online interest based and location based social network among mobile users. The two different networks are grouped into layers of a multilayer social network $M = \{G^{on}, G^{off}\}$, as shown in Figure 8. $G^{off}$ and $G^{on}$ depict offline and online social network separately. In each layer, the graph is described as $G = <$

V, E >, where V and E represent for node sets and edge sets., respectively. Nodes, such as $u_1, ..., u_4$, represent users. Edges exist among users when users share similar object-based interests [150]. Combining information from multiple networks in a multilayer configuration provides new insights into user interactions in online and offline world. It shed light on the link generation process from multiple views, which will significantly improve the link prediction systems with valuable applications to social bootstrapping and friend recommendations [149].

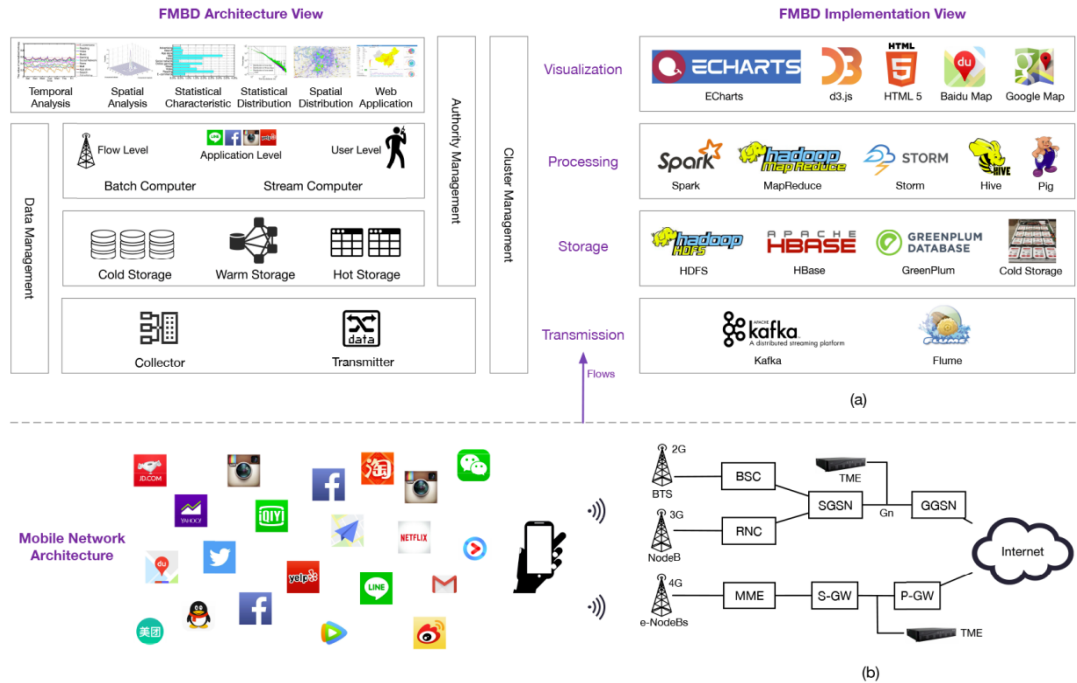

Fig. 7 The overall architecture of Framework for Mobile Big Data (FMBD) and our considered mobile networks architecture [60].

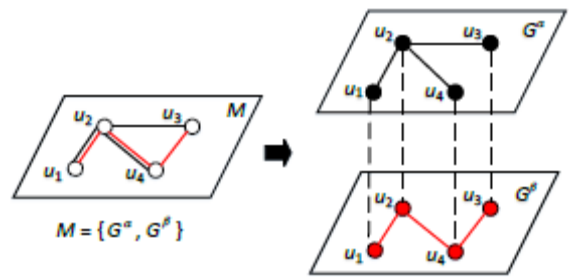

Fig. 8 Multilayer model of a network.

So far, we have summarized some representative works related to human online and offline behaviors. It is meaningful to note that due to the highly spatial-temporal and nonhomogeneous nature of the data traffic, a pervasive framework is challenging yet indispensable to realize the collection, processing and analyses of massive data, reducing resource consumption and improving Quality of Experience (QoE). The seminal work by Qiao et al. [60] proposes a framework for MBD, referred to as FMBD. It provides data collection, storage, processing, analyzing, and management functions to monitor and analyze the massive data traffic. Figure 7(a) displays the architecture of

FMBD, while Figure 7(b) shows the considered mobile networks framework. With the interaction between user equipment and 2G/3G/4G network, real massive mobile data can be collected by traffic monitoring equipment (TME). The implementation modules are based on Apache open-source software [151]. FMBD builds a security environment and easy-to-use platform both for operators and data analysts, showing good performance on energy efficiency, portability, extensibility, usability, security, and stability. In order to meet the increasing needs of traffic monitoring and analysis in the current and future mobile networks, it provides a solution to deal with large scale mobile big data.

In conclusion, the rapid development of mobile applications and increasing demands of accessing Internet by end users present both challenges and opportunities for the future mobile network. This section surveys the literature on analyses of human online and offline behavior based on the mobile traffic data. Moreover, a framework has also been investigated, in order to meet the higher requirement of dealing with dramatically increased mobile big data. The analyses of mobile big data will provide valuable information for the ISPs on network deployment, resource management, and the design of future mobile network architectures.

## 4. CONCLUSIONS AND FUTURE CHALLENGES

Although the machine learning-based methods introduced in Section 3 are widely applied in the MBD fields, and obtain good performances in real data test, the present methods still need to be further developed. Therefore, five main challenges facing MBD analysis regarding the machine learning-based methods should be considered as follow.

**1) Large-scale and high-speed M-Internet**

The growth of MIDs and high speed of M-Internet introduce massive and increasingly various mobile data traffic resulting in a heavy load to the wireless transmission system, which leads us to improve wireless communication technologies including WLAN and cellular mobile communication. In addition, the requirement of real-time services and applications demands on the development of machine learning-based MBD analysis methods towards high efficiency and precision.

**2) Overfitting and underfitting problems**

A benefit of MBD to machine learning and deep learning lies in the fact that the risk of overfitting becomes smaller with more and more data available for training [28]. However, underfitting is another problem for the oversize data volume. In this condition, a larger model might be a better selection, while the model can express more hidden information of the data. Nevertheless, larger model which generally implies a deeper structure increases runtime of the model which affects the real-time performance. Therefore, the model size in machine learning and deep learning, which represents number of parameters, should be balanced to model performance and runtime.

**3) Generalization problem**

As the massive scale of MBD, it is impossible to gain entire data even if they are only in a specific field. Therefore, the generalization ability which can be defined as suitable of different data subspace, or called scalability, of a trained machine learning or deep learning model is of great importance for evaluating the performance.

**4) Cross-modal learning**

The variety of MBD causes multiple modalities of data (for example, images, audios, personal location, web documents, and temperature) generated from multiple sensors (correspondingly, cameras, sound recorders, position sensor, and temperature sensor). Multimodal learning should learn from multiple modalities and heterogeneous input signals with machine learning and deep learning [4], [152], and obtain hidden knowledge and meaningful patterns, however, it is quite difficult to discover.

### 5) Extended channel dimensions

The channel dimensions have been continuously expanded from time domain to space, time and frequency three domains, which mean that the channel properties are deeply mined. Meanwhile, the increased antenna number, huge bandwidth and versatile application scenarios bring the big data of channel measurement, especially for 5G. The finding channel characteristics need to be precisely described by more advanced channel modeling methodologies.

In this paper, the applications and challenges of machine learning-based MBD analysis in the M-Internet have been reviewed and discussed. The development of MBD requires advanced data analysis technologies especially machine learning-based methods. Three typical applications of MBD analysis focus on wireless channel modeling, human online and offline behavior analysis, and speech recognition and verification in the Internet of vehicles, respectively, and the machine learning-based methods used are widely applied in many other fields. In order to meet the aforementioned future challenges, three main study aims, i.e. accuracy, feasibility and scalability [28], are highlighted for present and future MBD analysis research. In future work, accuracy improving will be also the primary task on the basis of a feasible architecture for MBD analysis. In addition, as the aforementioned discussion of the generalization problem, scalability has obtained more and more attentions especially in a classification or recognition problem where scalability also includes the increase in the number of inferred classes. It is of great importance to improve the scalability of the methods with the high accuracy and feasibility in order to face the analysis requirements of MBD.

**ACKNOWLEDGMENTS**
This paper is supported by …